\pdfoutput=1

\documentclass[11pt]{article}

\usepackage[preprint]{acl}

\usepackage{times}
\usepackage{latexsym}

\usepackage[T1]{fontenc}

\usepackage[utf8]{inputenc}

\usepackage{microtype}

\usepackage{inconsolata}

\usepackage{algorithm}
\usepackage{algorithmic} 
\usepackage{graphicx}    
\usepackage{csquotes}    
\usepackage{subcaption}  
\usepackage{ragged2e}    
\usepackage{enumitem}
\setlist{nolistsep}
\usepackage{multirow}
\usepackage{array}
\newcolumntype{A}{>{\centering\arraybackslash}m{0.15\linewidth}}
\newcolumntype{B}{>{\centering\arraybackslash}m{0.20\linewidth}}
\newcolumntype{C}{>{\centering\arraybackslash}m{0.30\linewidth}}
\newcolumntype{X}{>{\centering\arraybackslash}m{0.38\linewidth}}
\newcolumntype{M}{>{\centering\arraybackslash}m{0.25\linewidth}}
\newcolumntype{N}{>{\centering\arraybackslash}m{0.65\linewidth}}

\newcommand*{\affaddr}[1]{#1}
\newcommand*{\affmark}[1][*]{\textsuperscript{#1}}

\title{Can Language Models Act as Knowledge Bases at Scale?}

\author{Qiyuan He\affmark[\dag]
        \hspace{3mm}
        Yizhong Wang\affmark[\ddag]
        \hspace{3mm}
        Wenya Wang\affmark[\dag]
        \hspace{3mm} \\
        \affaddr{\affmark[\dag]Nanyang Technological University}\\
        \affaddr{\affmark[\ddag]University of Washington}\\
        \affmark[\dag]qiyuan001@e.ntu.edu.sg, wangwy@ntu.edu.sg,
        \affmark[\ddag]yizhongw@cs.uw.edu}
  
\begin{document}
\maketitle

\begin{abstract}
Large language models (LLMs) have demonstrated remarkable proficiency in understanding and generating responses to complex queries through large-scale pre-training. However, the efficacy of these models in memorizing and reasoning among large-scale structured knowledge, especially world knowledge that explicitly covers abundant factual information remains questionable. Addressing this gap, our research investigates whether LLMs can effectively store, recall, and reason with knowledge on a large scale comparable to latest knowledge bases (KBs) such as Wikidata.
Specifically, we focus on three crucial aspects to study the viability: (1) the efficiency of LLMs with different sizes in memorizing the exact knowledge in the large-scale KB; (2) the flexibility of recalling the memorized knowledge in response to natural language queries; (3) the capability to infer new knowledge through reasoning. Our findings indicate that while LLMs hold promise as large-scale KBs capable of retrieving and responding with flexibility, enhancements in their reasoning capabilities are necessary to fully realize their potential\footnote{Our datasets and source code can be obtained from \url{https://github.com/hyanique/LMKB-at-Scale}}. 
\end{abstract}

\section{Introduction}

The access to knowledge is critical for language models (LMs) to perform well on many tasks and serve users reliably.
Existing studies have found that language models, after pre-training, can encode a large amount of factual knowledge as well as implicit linguistic knowledge from the general corpus, making them a crucial component for tasks that require natural language understanding \cite{bommasani2022opportunities,li-etal-2022-systematic}.
This leads to the potential of using language models as knowledge bases \cite{petroni-etal-2019-language, alkhamissi2022review}.
However, existing studies mainly focus on probing \cite{li-etal-2022-eliciting,chen-etal-2022-probing,sung-etal-2021-language} and utilizing \cite{roberts-etal-2020-much,moiseev-etal-2022-skill} LMs' knowledge gained from pre-training,
which shows deficiencies when handling long-tail, less frequently appeared knowledge \cite{kandpal2023large}, due to knowledge imbalance, conflict, and noise in the pre-trained corpora \cite{Carlini2022QuantifyingMA, razeghi-etal-2022-impact, tanzer-etal-2022-memorisation}.

Meanwhile, knowledge bases (KBs), commonly utilized in many knowledge-intensive tasks such as dialogue \cite{li-etal-2022-knowledge, galetzka-etal-2021-space}, question answering \cite{baek-etal-2023-knowledge, saxena-etal-2020-improving, Qiu2020StepwiseRF} and recommendation systems \cite{Wu2013OntologybasedSQ}, are known for their ability to compactly organize information on a large scale, providing clean and balanced knowledge. For example, Wikidata contains over 108M entities about the world\footnote{\url{https://www.wikidata.org/wiki/Wikidata:Statistics}}.
Operations over larger KBs lead to greater computational costs and therefore, pose a big challenge for extracting subgraphs from the KB \cite{Cordella2004AI, Grohe2020TheGI} or grounding semantic logic forms over the KB \cite{lan-jiang-2020-query, Bhutani2019LearningTA} to perform downstream tasks. In addition, the rigid format of KBs limit their flexibility to handle complex natural language queries.

In this work, we propose to explicitly train large language models to memorize world knowledge from Wikidata \cite{Tanon2016WikiData} at a large scale and systematically study the viability of using the resulting LMs to function as the knowledge base.
With their high capacity, we hypothesize that LMs can store information from a knowledge base on a rather large scale and provide more flexibility in querying and reasoning. Specifically, we aim to answer the following three questions: (1) How fast and how good can LMs with different sizes memorize large-scale knowledge of different frequencies through training? (2) How flexible are these trained LMs when being used to answer queries in natural languages rather than the structured triplets used during training? (3) Can LMs infer new knowledge that does not exist in the KB, and what kind of reasoning capabilities are involved?
We distinguish our work from those that train LMs on small-scale KBs with popular facts \cite{heinzerling-inui-2021-language} or convert knowledge triplets to synthetic sentences using manually curated templates \cite{heinzerling-inui-2021-language,petroni-etal-2019-language} which only works for a limited set of relations.

We start by proposing an efficient learning algorithm based on importance sampling \cite{Alain2015VarianceRI, katharopoulos2019samples, Zhang2019AutoAssistAF} to train LMs to memorize knowledge more efficiently. To answer the first question, we evaluate the memorization capacity of LMs of different sizes as well as their performances on both popular and long-tail world knowledge.
We observe that LMs are capable of memorizing information from a knowledge base on a large scale, with larger model learning faster. In addition, infrequent knowledge is more challenging to memorize, irrespective of the size of the language models. 

To answer the second question on LMs' flexibility in handling natural language queries, we further finetune the trained LMs using PopQA \cite{mallen-etal-2023-trust}, a natural language QA dataset that requires long-tail Wikidata knowledge. With minimal finetuning, these LMs demonstrate superior performance over their counterpart, which are not trained on Wikidata KB. This indicates the power of LMs in flexibly retrieving and organizing long-tail knowledge, regardless of the presentation form, unveiling their potential for responding to various user queries.

To answer the third question from the perspective of incomplete KBs, we use the dataset released by \citet{veseli-etal-2023-evaluating} containing general missing facts (triplets) and further curate two sets of missing facts tailoring two kinds of reasoning capabilities, namely inverse reasoning by switching the positions of the subject and object, and compositional reasoning which conjoins two relations to form a new one. By evaluating LMs' performances on inferring the missing facts, we study their inherent reasoning capabilities in addition to memorizing existing facts.
Our results show that LMs are capable of inferring missing entities from existing knowledge to some extent via reasoning. However, they struggle with inverse reasoning more often than compositional reasoning, advocating for further investigations and explorations on how to improve LMs' reasoning capabilities, specifically, inverse reasoning. 

\section{Training LMs on Large-Scale KB}

\subsection{KB Dataset}
A basic knowledge base is a collection of facts in the form of \textit{(subject, relation, object)} triplets, for example, Freebase \cite{Bollacker2008FreebaseKB} and DBPedia \cite{Auer2007DBpediaAN}. To study the memorization capacity of language models at a large scale, we consider Wikidata \cite{Tanon2016WikiData}, one of the largest knowledge bases to date that is actively maintained by the community. Compared with pre-training corpora, Wikidata contains abundant world knowledge in a more compact and accurate form, covering both popular and long-tail knowledge that appears less frequently in the pre-training corpora of LMs. 

\begin{figure}[ht!]
\captionsetup[subfigure]{}
\begin{subfigure}[t]{0.47\linewidth}
    \includegraphics[width=\linewidth]{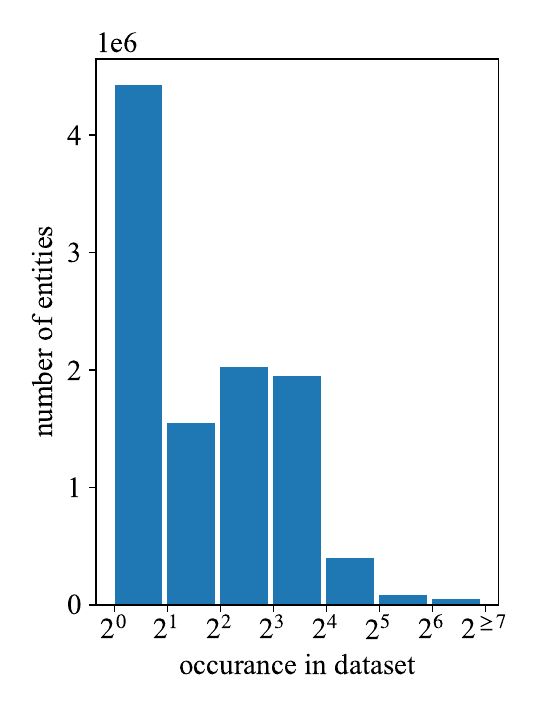}
    \caption{Distribution of entities in $\mathcal{D}_0$, with number of entities in 1e6 scale and occurrence in the powers of 2. }
    \label{fig:46M-distro-ent}
\end{subfigure}
\hspace{\fill}
\begin{subfigure}[t]{0.47\linewidth}
    \includegraphics[width=\linewidth]{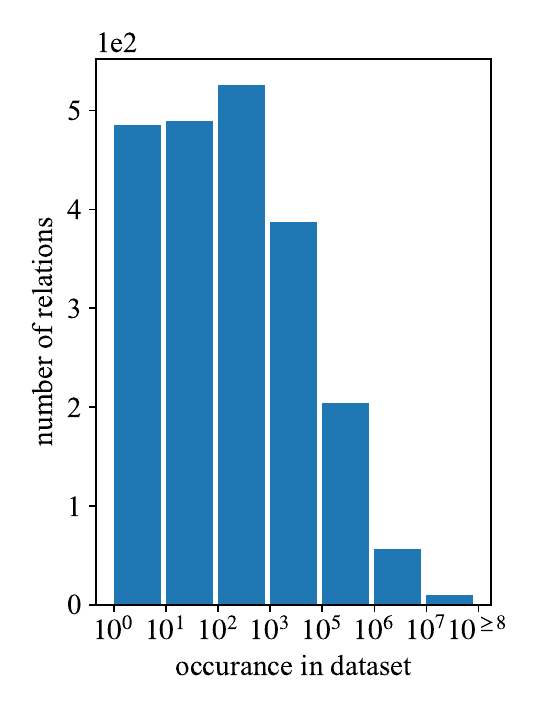}
    \caption{Distribution of relations in $\mathcal{D}_0$, with occurrence count in the powers of 10.}
    \label{fig:46M-distro-rel}
\end{subfigure}
\caption{Distribution of entity and relation occurrences in world knowledge $\mathcal{D}_0$.}
\label{fig:46M-distro}
\end{figure}

When preparing the KB dataset, we use the cleaned knowledge taken from the 2022 January snapshot of Wikidata dump \cite{philipp2021wikidatacore} to avoid knowledge irrelevant to common question-and-answer, specifically, filtering away URLs, images, geographical coordinates, and subject entities that do not have a corresponding Wikipedia page. If there are multiple objects given a subject and a relation, we randomly sample a single instance from the available objects to avoid knowledge ambiguity.
After filtering, we obtain a dataset of 46M \textit{(subject, relation, object)} triplets, with the distribution of 10.5M entities (subjects or objects) and 2,157 relations shown in Figure~\ref{fig:46M-distro-ent} and Figure~\ref{fig:46M-distro-rel}. We denote this dataset as $\mathcal{D}_0$. We can observe that over 4M entities only appear once or twice, and around 500 relations appear 1-10 times. Meanwhile, around 250 relations occur more than 10K times, and 530K entities make more than 16 occurrences. These statistics show that $\mathcal{D}_0$ covers adequate popular knowledge as well as a non-neglectable portion of long-tail knowledge. 

To study how the model performs regarding knowledge frequency inside the KB, we first calculate the number of occurrences of all entities and relations. Next, we define \emph{long-tail} entities/relations as entities/relations of top 15\% when ranking all entities/relations by their numbers of occurrences in ascending order and \emph{popular} entities/relations as entities/relations of top 5\% when ranking them by their numbers of occurrences in descending order. Then, under each of the \emph{long-tail} and \emph{popular} categories, we randomly sample triplets under both the entity set and the relation set, resulting in four datasets denoted as $\mathcal{D}_{PopRel}$, $\mathcal{D}_{PopEnt}$, $\mathcal{D}_{TailRel}$, and $\mathcal{D}_{TailEnt}$. As $\mathcal{D}_0$ contains 2,157 relations, the number of knowledge with \emph{long-tail} relations is limited\footnote{$\mathcal{D}_0$ contains 323 long-tail relations that occur 1-2 times in $\mathcal{D}_0$, summed to 663 occurrences in total. In comparison, the top $5\%$ of 2,157 relations make 40.8M occurrences in  $\mathcal{D}_0$}, leading to 663 samples in $\mathcal{D}_{TailRel}$. The other three datasets contain 1K triplets each. Example triplets include (``Linlithgow Burgh Halls'', \textit{instance of}, ``Town hall'') from $\mathcal{D}_{PopRel}$ and (``Department of Agriculture, Water and the Environment'', \textit{external auditor}, ``Australian National Audit Office'') from
 $\mathcal{D}_{TailRel}$.

\subsection{Model Setup}

We choose two language models, namely T5 \cite{Raffel2019ExploringTL} and LLaMA-2 \cite{Touvron2023Llama2O}, each with two different sizes: T5-base, T5-large, LLaMA-2-7b, and LLaMA-2-13b.
Starting from their pre-trained checkpoints, we continue training these models on the filtered Wikidata KB $\mathcal{D}_0$ containing 46M knowledge triplets. 
See Appendix~\ref{sec:appx-model} for the detailed training setup.

For each knowledge triplet in the form of \textit{(subject, relation, object)}, we create an input string by concatenating the prefix ``Subject:'' followed by the \textit{subject} text, the prefix ``Relation:'' followed by the \textit{relation} text and the prefix ``Object:'', and use the \textit{object} text as the output. 
For example, given the knowledge triplet (``Palaeontological Museum, Munich'', \textit{architect}, ``Leonhard Romeis''), the input to the LMs is ``\underline{Subject}: Palaeontological Museum, Munich. \underline{Relation}: architect. \underline{Object}:'' and the expected output is the object ``Leonhard Romeis''.

The training objective is to maximize the probability of generating the correct object: $p_{LM}(x_{out}|x_{in})$ where $x_{out}$ is the object text and $x_{in}$ is the input text. $p_{LM}$ denotes the probability distribution given by the language model.
\subsection{Importance Sampling}\label{sec:imsmp}

With the goal of injecting abundant and diverse information from large-scale KB information into LMs, it is imperative for the model to converge to a state where it can, in an ideal scenario, memorize every triplet within the training dataset. Traditional training process iterates through each data sample precisely once during each epoch, inherently treating all data with uniform importance. This approach, however, leads to extended training durations and reduced convergence efficiency, particularly when dealing with large-scale KBs containing a significant amount of hard-to-memorize knowledge. To address this issue, inspired by the importance sampling algorithm proposed in \cite{Alain2015VarianceRI,katharopoulos2019samples}, we allocate distinct importance weights to the training samples within $\mathcal{D}_0$. The importance weight is proportional to the prediction loss of each sample, serving as a measure of its memorization difficulty. This strategy prioritizes samples that are more challenging to memorize by assigning them greater importance, thereby increasing their likelihood of selection during each training iteration, leading to faster convergence speed \cite{Zhang2019AutoAssistAF, Xie2023DataSF}. 

The detailed algorithm is shown in Algorithm~\ref{algo: km-w-imsmp}.
\begin{algorithm}[htp]
 \caption{Knowledge memorization with importance sampling}
 \label{algo: km-w-imsmp}
    \begin{algorithmic}[1]
 \REQUIRE knowledge samples with importance $\mathcal{D}=\left\{(x_1, y_1; w_1), ..., (x_n, y_n; w_n)\right\}$
 \REQUIRE language model pre-trained on general corpora
 \ENSURE sampling ratio $\alpha \in (0,1)$
 \STATE initialize importance $w_1,...,w_n$ with $1e6$
 \FOR{every training epoch $e$}
  \STATE sample $\mathcal{S}=\left\{(x^s, y^s; w^s)\right\}\subset\mathcal{D}$ of size $n\times\alpha$ using importance $w_1,...,w_n$
  \STATE forward pass using $\left\{(x^s, y^s)\right\}$
  \STATE update importance $w^s$ using instance loss $\mathcal{L}(y^s,x^s)$
  \STATE backpropagation
  \ENDFOR
 \end{algorithmic}
\end{algorithm}
As shown in the pseudo-code, we use instance loss $\mathcal{L}(y^s,x^s)$  to measure the knowledge triplet’s importance and use this importance as the sampling probability within each batch, where $\mathcal{L}$ is the cross-entropy loss, and $y^s$ is the correct output text given input $x^s$. Mathematically, 
\begin{equation}
\mathcal{L}(y^s,x^s)=-\sum_{t=1}^T \log{p_{LM}(y_t^s|x^s)},
\end{equation}
with $T$ being the number of tokens in $y^s$ and $y_t^s$ being the $t$-th token in $y^s$,
Hence, the higher the instance loss, the higher the chance for the instance to be sampled into the subset $\mathcal{S}$ for training, forcing the model to focus on learning hard samples more often.

\begin{figure}[ht!]
    \centering
    \includegraphics[width=0.99\linewidth]{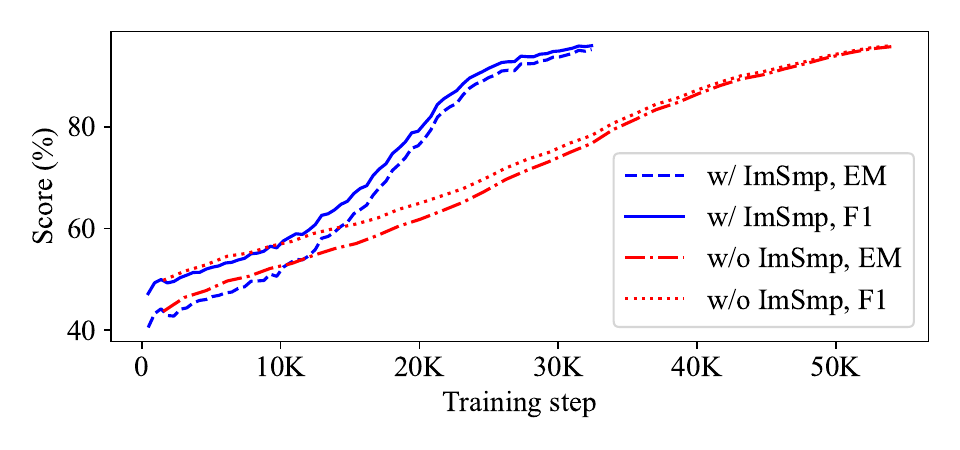}
    \caption{Learning curves of T5-base training on $\mathcal{D}_{1}$ with and without importance sampling (ImSmp), evaluated using $\mathcal{D}_{1-Eval}$.}
    \label{fig: ImSmp-Curve}
\end{figure}
To verify our hypothesis, we conduct a preliminary experiment by randomly sampling 1\% triplets from $\mathcal{D}_0$ and train a T5-base model to memorize this sampled dataset, with and without using Algorithm~\ref{algo: km-w-imsmp}. We denote this subset containing 426K triplets as $\mathcal{D}_1$. We further arbitrarily sampled 10K triplets from $\mathcal{D}_1$ as the corresponding evaluation set, denoted as $\mathcal{D}_{1-eval}$. We configure the sampling ratio $\alpha$ to be 0.3.
As shown in Figure~\ref{fig: ImSmp-Curve}, the model trained without importance sampling quickly reaches around 80\% exact match and F1 score in the first 30K training steps, and then its performance slowly increases to around 95\% exact match and F1 score using another 20K steps. But with importance sampling, the model achieved roughly $80\%$ exact match and F1 score after the first 20K steps, and over 95\% exact match and F1 score after another 12K steps. We also notice that training with importance sampling yields a significantly steeper learning curve when compared with the one without importance sampling. In what follows, we stick with importance sampling with the same $\alpha$ value when training LMs for all the experiments.

\subsection{Evaluation}\label{sec:eval-setup}

To study the LM's capacity of memorizing the structured knowledge base, we propose to use the exact match (EM) and F1 scores following \cite{heinzerling-inui-2021-language} over the entire training dataset. We call this \textbf{fixed-form information recall} ability. Since it is not feasible to iteratively evaluate the LM on all 46M triplets in $\mathcal{D}_0$ throughout the training process due to huge inference time, we opt to randomly sample 10K triplets from $\mathcal{D}_0$ as the evaluation set, denoted as $\mathcal{D}_2$.

\begin{figure*}[htp]
\captionsetup[subfigure]{}
\begin{subfigure}[t]{0.32\textwidth}
    \includegraphics[width=\textwidth]{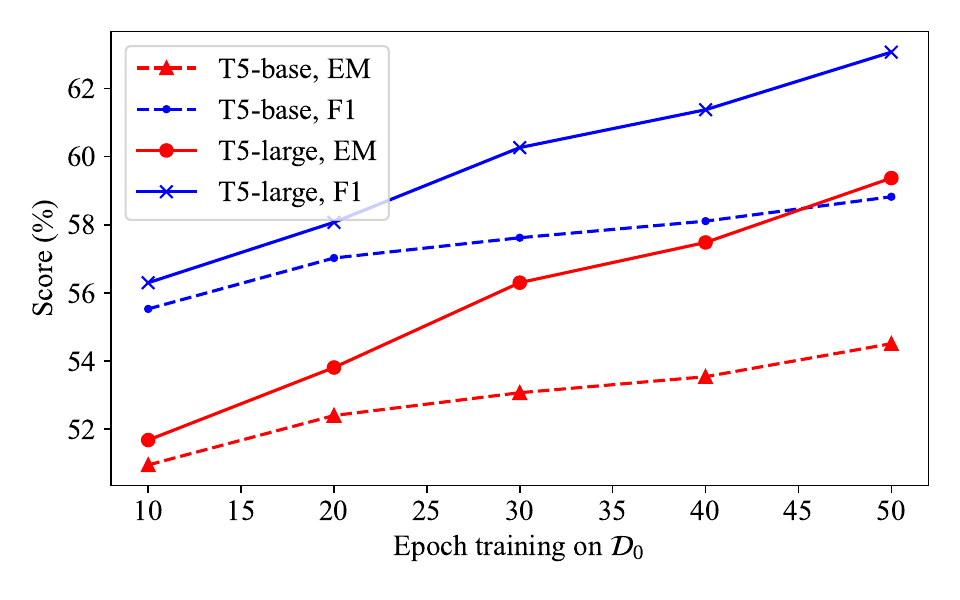}
    \caption{Evaluating T5 performance using $\mathcal{D}_2$.}
    \label{fig:T5-kmem}
\end{subfigure}
\hspace{\fill} 
\begin{subfigure}[t]{0.32\textwidth}
    \includegraphics[width=\linewidth]{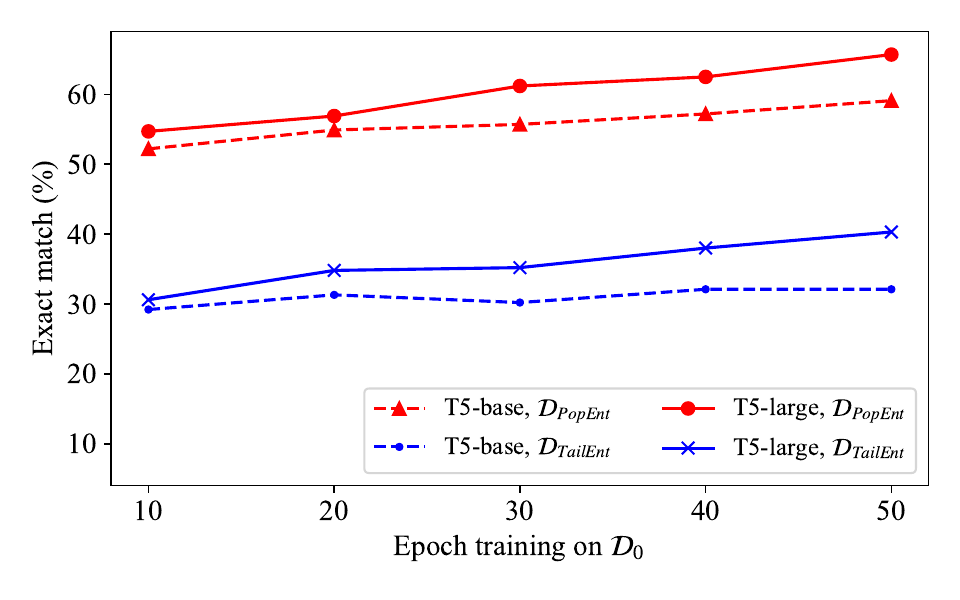}
    \caption{Evaluating T5 performance using $\mathcal{D}_{PopEnt}$ and $\mathcal{D}_{TailEnt}$.}
    \label{fig:T5-Eval-ent}
\end{subfigure}
\hspace{\fill} 
\begin{subfigure}[t]{0.32\textwidth}
    \includegraphics[width=\linewidth]{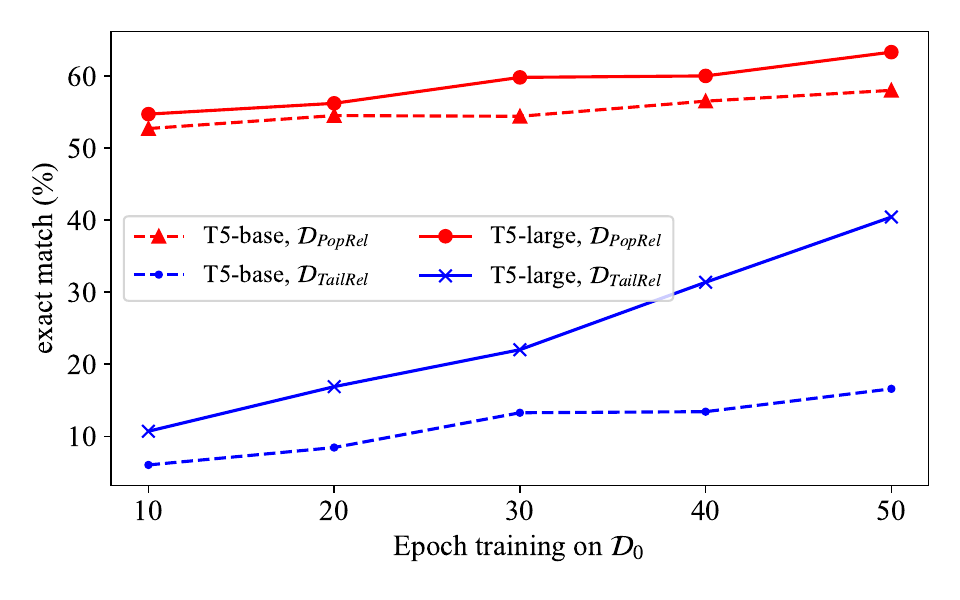}
    \caption{Evaluating T5 performance using $\mathcal{D}_{PopRel}$ and $\mathcal{D}_{TailRel}$.}
    \label{fig:T5-Eval-rel}
\end{subfigure}
\bigskip 
\begin{subfigure}[t]{0.32\textwidth}
    \includegraphics[width=\linewidth]{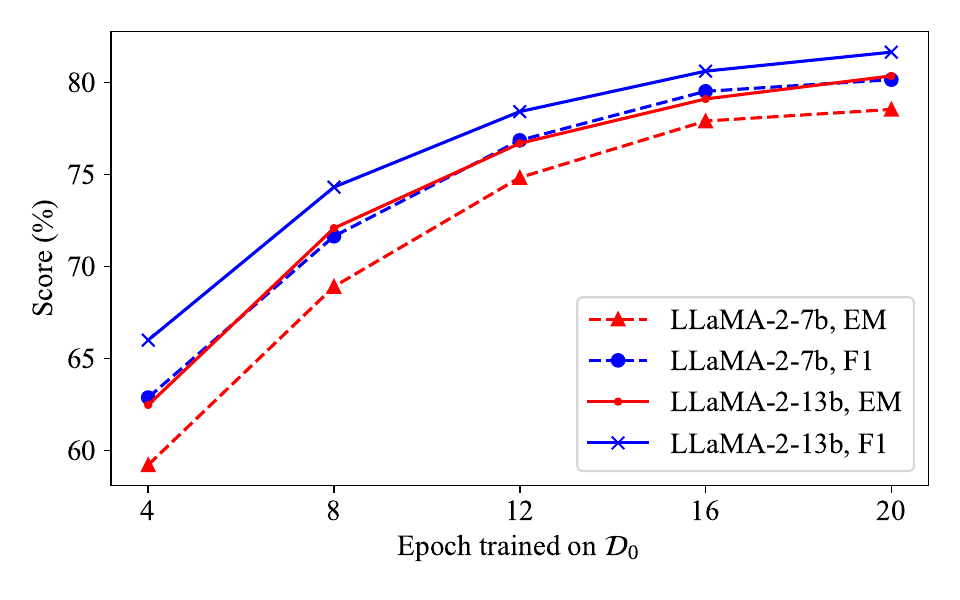}
    \caption{Evaluating LLaMA-2 performance using $\mathcal{D}_2$.}
    \label{fig:Llama-kmem}
\end{subfigure}
\hspace{\fill} 
\begin{subfigure}[t]{0.32\textwidth}
    \includegraphics[width=\linewidth]{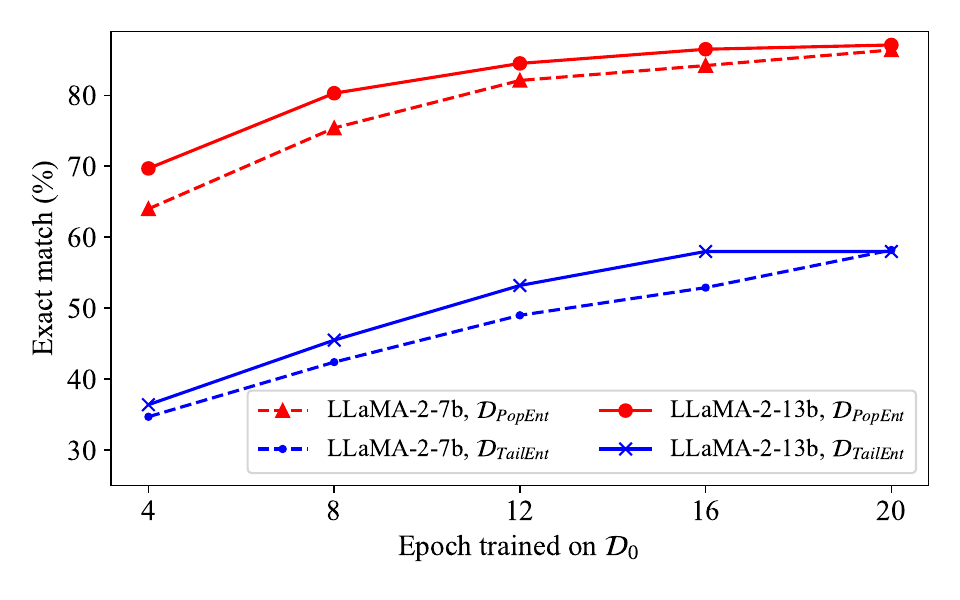}
    \caption{Evaluating LLaMA-2 performance using $\mathcal{D}_{PopEnt}$ and $\mathcal{D}_{TailEnt}$.}
    \label{fig:Llama-Eval-ent}
\end{subfigure}
\hspace{\fill} 
\begin{subfigure}[t]{0.32\textwidth}
    \includegraphics[width=\linewidth]{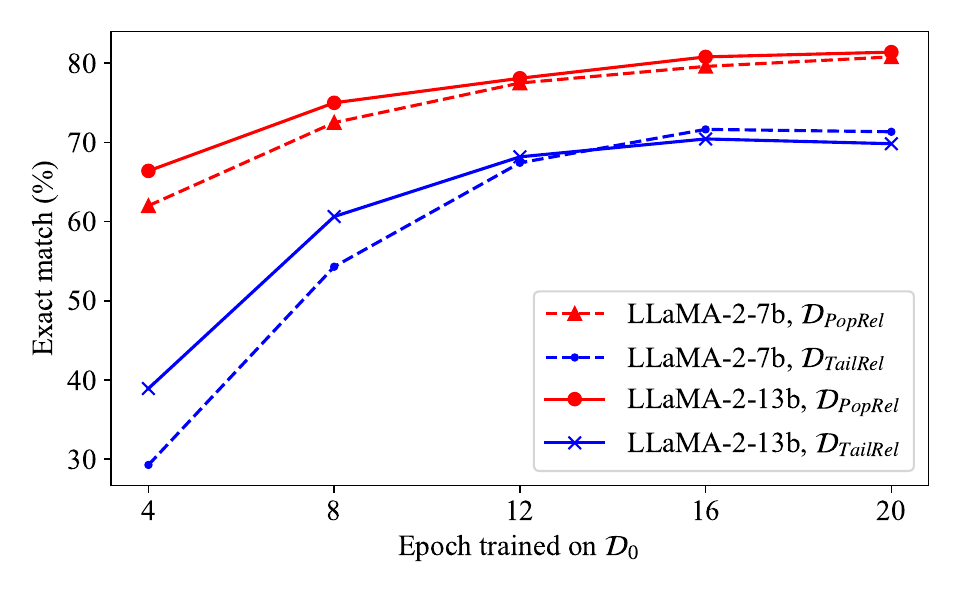}
    \caption{Evaluating LLaMA-2 performance using $\mathcal{D}_{PopRel}$ and $\mathcal{D}_{TailRel}$.}
    \label{fig:Llama-Eval-rel}
\end{subfigure}
\caption{Evaluating the fixed-form information recall ability for LMs training on $\mathcal{D}_0$. T5 models are on the upper row, and LLaMA-2 models are on the bottom row.}
\end{figure*}

To measure the model's ability to flexibly retrieve memorized knowledge when queried with an input and output format different from training, we consider using natural language to query our model the same way as the task of question answering (QA). We call this \textbf{free-form information recall} ability. For implementation, we require that the knowledge used by the QA task should be highly covered by the 46M triplets of the world knowledge from Wikidata. Hence, we select the QA dataset constructed in PopQA \cite{mallen-etal-2023-trust}. PopQA converted 14K knowledge triplets from Wikidata to their corresponding natural language questions and answers that cover long-tail information based on Wikipedia page views. With a random 8:2 split to obtain a train set of 11.4K samples and a validation set of 2.9K samples, we further finetune the model from the memorization checkpoints using the training split of PopQA and evaluate the performance on the validation set using the F1 score. We also compute the exact match and F1 score of the model's generation accuracy over the PopQA knowledge triplet to check if the model can access relevant knowledge using fixed-form recall.

Lastly, we explore whether LMs can infer new knowledge that does not exist in the KB, namely, the \textbf{missing fact completion} ability. Since most knowledge graphs are incomplete, missing factual triplets or even entities \cite{Yang2022RethinkingKG, Shi2017OpenWorldKG}, the ability to automatically complete missing facts becomes especially demanding.  
First we consider the missing facts dataset released by \citet{veseli2023evaluating}, which contains 350 factual triplets missing from Wikidata with human annotated ground-truths. 
As we additionally seek to investigate the underlying reasoning capabilities involved in missing fact completion, we also curate two sets of missing knowledge triplets based on $\mathcal{D}_0$, emphasizing \textbf{inverse reasoning} and \textbf{compositional reasoning}, respectively.
For a missing knowledge triplet that is not contained in $\mathcal{D}_0$, we query the model using the same input format as in fixed-form information recall and evaluate the output text against object text using F1 scores\footnote{For pre-trained models without training on knowledge base $\mathcal{D}_0$, we query the models with natural language inputs released alongside the triplets. See respective task in Section~\ref{sec:missing-facts} for details}.

Next, we present the detailed evaluation and analysis to answer each of the three core questions, including (1) the efficiency of LMs with different sizes in memorizing the exact knowledge in the large-scale KB (Section \ref{sec:KB}); (2) the flexibility of recalling the memorized knowledge in response to natural language queries (Section \ref{sec:PQA}); (3) the capability to infer new knowledge through reasoning (Section \ref{sec:reasoning}).

\section{Fixed-Form Information Recall} \label{sec:KB}

As mentioned in Section \ref{sec:eval-setup}, we measure the fixed-form information recall ability on a sub-sampled dataset $\mathcal{D}_2$ from the original training set $\mathcal{D}_0$ to avoid the huge inference cost. See Appendix~\ref{sec:appx-model} for additional training details. Specifically, we compute the exact match and F1 score on $\mathcal{D}_2$ along the training steps of T5-base, T5-large, LLaMA-2-7b and LLaMA-2-13b. The performance curves are shown in Figure~\ref{fig:T5-kmem} and \ref{fig:Llama-kmem}.

The results show that the models can memorize a large portion of 46M world knowledge, with T5-large performing better than T5-base, and LLaMA-2-13b slightly more capable than LLaMA-2-7b in terms of memorization capacity. LMs with larger sizes are capable of memorizing more knowledge with higher efficiency. In particular, at the end of training, LLaMA-2-13b gives the highest F1 score of 81.64\%, whereas T5-large reaches an F1 score of 63.07\%.

In addition, we further separately evaluate the performances on popular and long-tail triplets, i.e., $\mathcal{D}_{PopEnt}$, $\mathcal{D}_{PopRel}$, $\mathcal{D}_{TailEnt}$ and $\mathcal{D}_{TailRel}$. The results are shown in Figure~\ref{fig:T5-Eval-ent}, \ref{fig:T5-Eval-rel}, \ref{fig:Llama-Eval-ent} and \ref{fig:Llama-Eval-rel}. These plots demonstrate that
(1) All models are better at memorizing popular information than memorizing long-tail information;
(2) For LLaMA-2 models, a larger model size does not lead to significantly better memorization capability when it comes to long-tail and popular knowledge;
(3) Different from LLaMA-2, we observe that T5-large is better than T5-base in learning both popular and long-tail knowledge, with an even significant improvement for long-tail relations ($\mathcal{D}_{TailRel}$).

\section{Free-Form Information Recall}\label{sec:PQA}

To evaluate the model's ability to perform free-form information recall when using natural language queries, as indicated in Section \ref{sec:eval-setup}, we adopt the knowledge triplets and their corresponding natural language questions from PopQA: 

Given a knowledge triplet \textit{(``Binary'', \textit{author}, ``Michael Crichton'')} from Wikidata, PopQA converts it to a natural language question which asks for the object: ``\textit{Who is the author of Binary?}''. The correct answer in this case should be ``\textit{Michael Crichton}''. To make LMs trained on knowledge triplets familiar with the natural language QA format, we further finetune these LMs by feeding them the question as input and training these models to generate the correct answer. For T5, the input is the original question such as ``\textit{Who is the author of Binary?}''. For LLaMA-2, the input is ``\textit{\underline{Question}: Who is the author of Binary? \underline{Answer}:}''. We then evaluate the generated output using the F1 score\footnote{We use F1 score to allow minor linguistic variations in the generated output, taking into account semantic similarity and flexibility.}. In addition to the free-form queries, we also evaluate how much of the PopQA knowledge in its original triplet form is memorized by the model at each specific checkpoint by querying the model using the subject and relation, following the same input format used for fixed-form information recall.

\begin{figure}[ht!]
\begin{subfigure}[t]{0.45\textwidth}
\includegraphics[width=\linewidth]{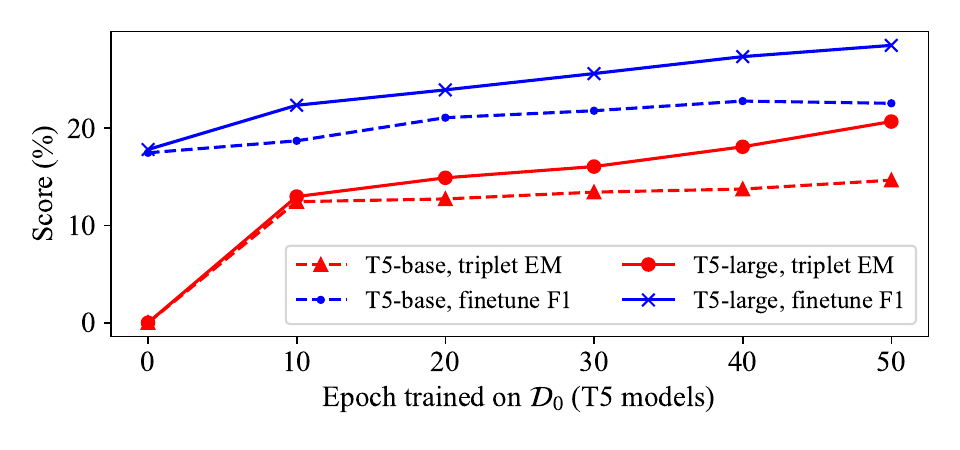}
\end{subfigure}
\begin{subfigure}[t]{0.45\textwidth}
\includegraphics[width=\linewidth]{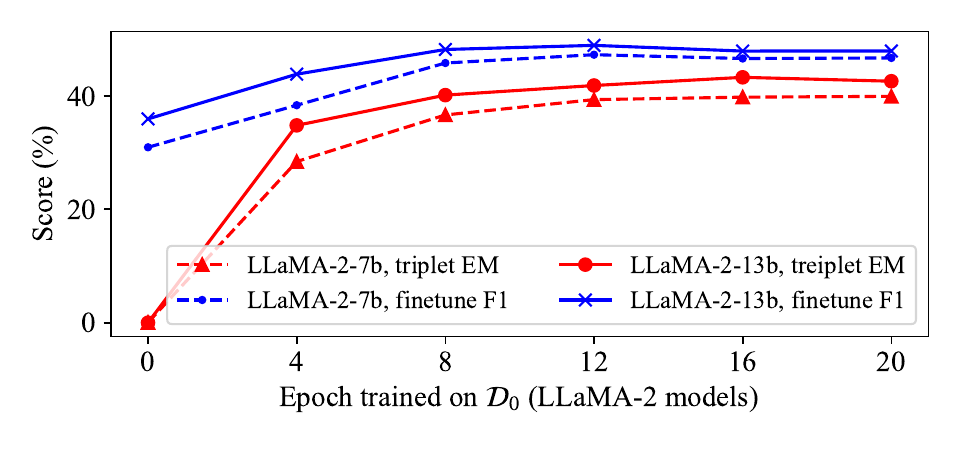}
\end{subfigure}
\caption{PopQA finetuning performance and knowledge recall on various checkpoints through training on $\mathcal{D}_0$. The pre-trained models are represented by epoch $0$.}
\label{fig: PQA-results}
\end{figure}

The results on PopQA are shown in Figure~\ref{fig: PQA-results}. 
Each point in the x-axis indicates the number of epochs for each checkpoint when training LMs using the Wikidata triplets, i.e., $\mathcal{D}_0$. Starting from each of these checkpoints, we further finetune the LMs using the training data from PopQA for up to 30 epochs for T5 models and 15 epochs for LLaMA-2 models with early stopping (see Appendix~\ref{sec:appx-model} for details) and report the best F1 score on the evaluation set.

It is clear that training on $\mathcal{D}_0$ can provide a sizable performance boost compared with using the originally pre-trained LMs (epoch=0). This suggests that LMs trained on large-scale knowledge bases are capable of performing some extent of free-form information recall, especially for a question-answering task that emphasizes long-tail knowledge. 
We also notice that memorizing more knowledge (as indicated by the triplet EM scores) leads to better performance in general.
In addition, larger models, after being trained on $\mathcal{D}_0$, are able to recall more knowledge for this downstream task in a fixed form, and finetuning yields better results. 

\section{Missing Fact Completion}\label{sec:reasoning}

\begin{figure*}[htp]
\captionsetup[subfigure]{}
\begin{subfigure}[t]{0.32\textwidth}
    \includegraphics[width=\textwidth]{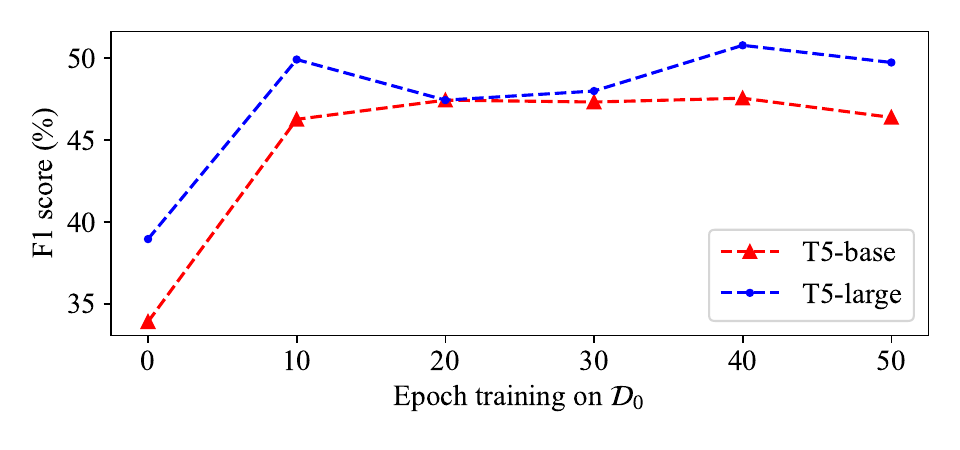}
    \caption{T5: general missing facts.}
    \label{fig:T5-kbc}
\end{subfigure}
\hspace{\fill} 
\begin{subfigure}[t]{0.32\textwidth}
    \includegraphics[width=\linewidth]{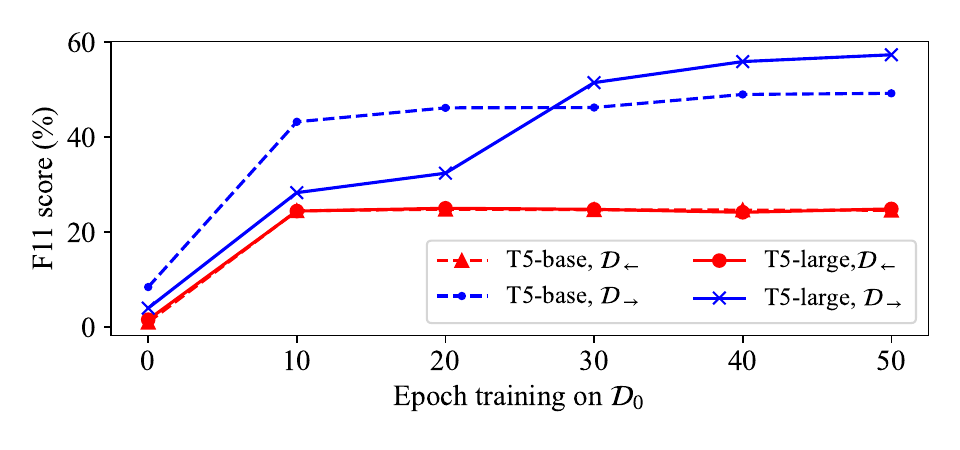}
    \caption{T5: inverse reasoning.}
    \label{fig:T5-onehop}
\end{subfigure}
\hspace{\fill} 
\begin{subfigure}[t]{0.32\textwidth}
    \includegraphics[width=\linewidth]{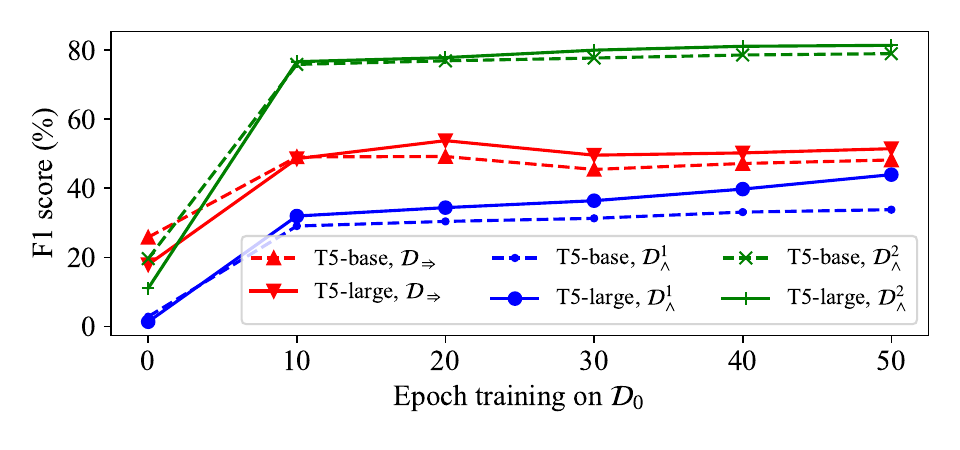}
    \caption{T5: compositional reasoning.}
    \label{fig:t5-twohop}
\end{subfigure}
\bigskip 
\begin{subfigure}[t]{0.32\textwidth}
    \includegraphics[width=\linewidth]{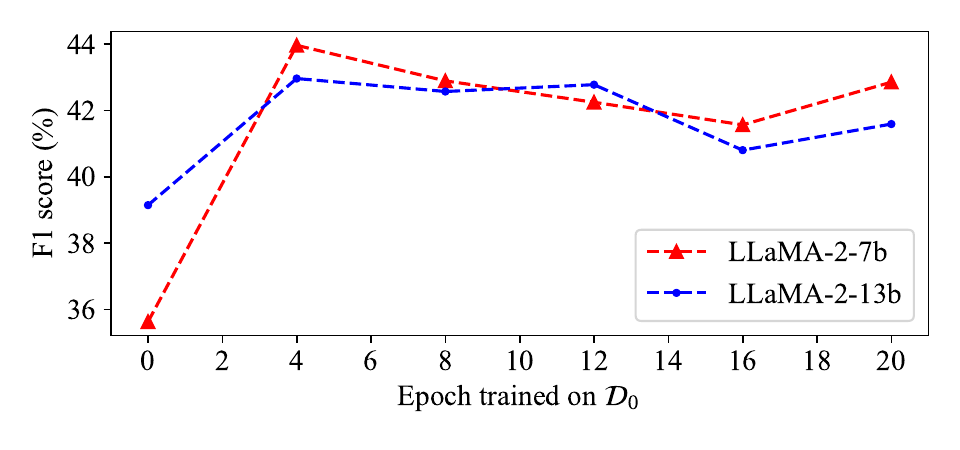}
    \caption{LLaMA-2: general missing facts.}
    \label{fig:llama-kbc}
\end{subfigure}
\hspace{\fill} 
\begin{subfigure}[t]{0.32\textwidth}
    \includegraphics[width=\linewidth]{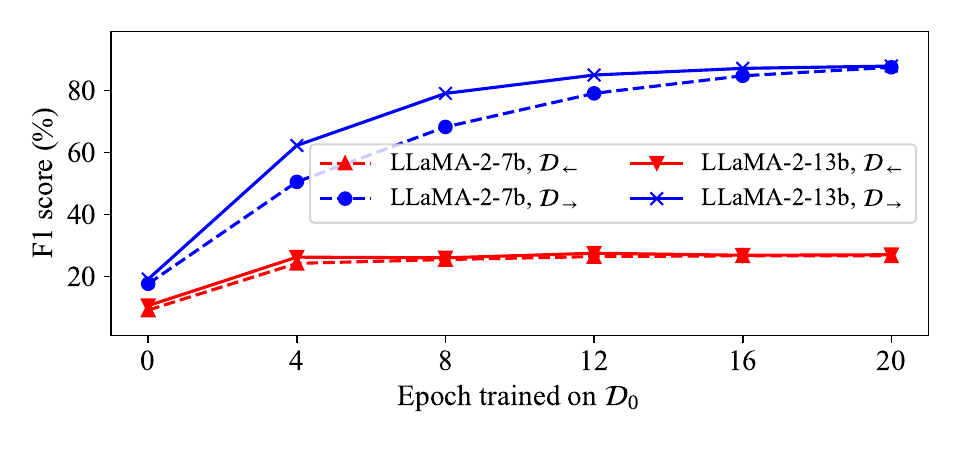}
    \caption{LLaMA-2: inverse reasoning.}
    \label{fig:llama-onehop}
\end{subfigure}
\hspace{\fill} 
\begin{subfigure}[t]{0.32\textwidth}
    \includegraphics[width=\linewidth]{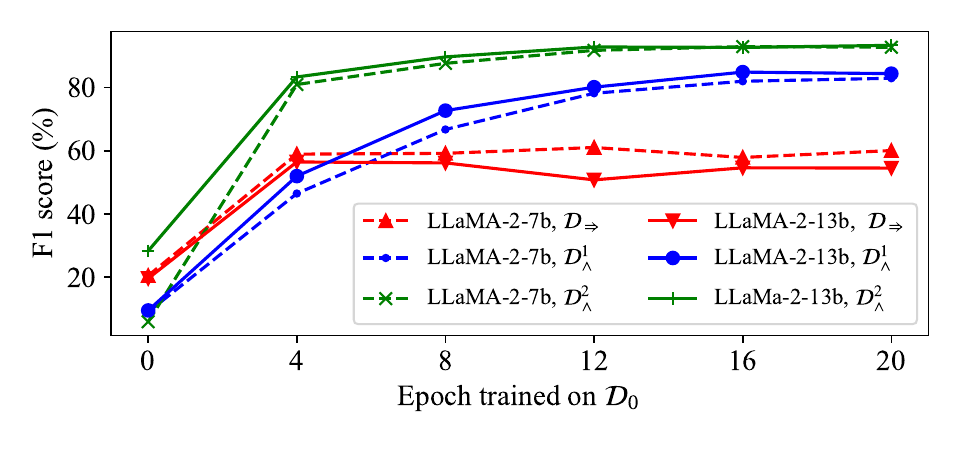}
    \caption{LLaMA-2: compositional reasoning.}
    \label{fig:llama-twohop}
\end{subfigure}
\caption{Evaluating the ability to infer new knowledge across various model checkpoints through training on $\mathcal{D}_0$. The x-axis of the plots indicates the checkpoints having the number of epochs when training LMs using $\mathcal{D}_0$. Specifically, epoch $0$ stands for the pre-trained checkpoints.}
\end{figure*}

\subsection{General Missing Facts} \label{sec:missing-facts}

To evaluate how the model performs when completing missing facts in general, we consider knowledge triplets that are missing from $\mathcal{D}_0$. We query the model to generate an object given the subject and relation. To ensure the feasibility of this setting, we require that the subject and relation in question are both contained in the knowledge base. Hence, the model has to associate relevant information related to the subject and the relation in order to infer the object.

For implementation, we utilize the missing fact dataset \cite{veseli2023evaluating} consisting of 350 samples of knowledge missing from Wikidata. For each sample, we query the model using the subject and the relation that are contained in Wikidata, and compare the generated output with the human-annotated object using the F1 score\footnote{When there are multiple ground-truth candidates, we compare the generated results to each of them and take the best F1 scores}. To clearly demonstrate the benefit of knowledge memorization, we further evaluate how the pre-trained LMs perform on these missing facts using the natural language queries 
provided by the dataset. For example, a missing fact triplet (``Tidö Castle'', \textit{headquarters location}, ``Västeras'') is associated with the following natural language question ``The headquarter of Tidö Castle is in'' as input for T5, while the input for LLaMA-2 is ``\underline{Question}: The headquarter of Tidö Castle is in? \underline{Answer}:''.

As shown in Figure~\ref{fig:T5-kbc} and \ref{fig:llama-kbc}, we can see that training on $\mathcal{D}_0$ provides some performance increase. This suggests that training on large-scale knowledge bases can help LMs to infer new facts better. Furthermore, the capability of LMs to infer new facts does not grow along with the memorization process on $\mathcal{D}_0$ and larger models like LLaMA-2 even perform worse than smaller models like T5. These observations indicate that the amount of knowledge learned by the models may not be the key factor to determine their inference capability towards missing facts.

\subsection{Inverse Reasoning}\label{sec:onehop}

We define inverse reasoning as the ability to infer $(B,r',A)$ given the triplet $(A,r,B)$, where $A$ and $B$ represent two entities and $r,r'$ indicate relations. To study the model’s ability to conduct inverse reasoning over the knowledge base, we first curate a set of triplets in the form of $(A,r,B)$ originally contained in $\mathcal{D}_0$, denoted as $\mathcal{D}_{\rightarrow}$. Then, we curate the inverse set by mapping the original relation $r$ to its inverse $r'$ and switch the positions of $A$ and $B$, forming the triplets $(B, r', A)$. We denote this set using $\mathcal{D}_{\leftarrow}$. We query the model for the object entity $A$ when given the subject entity $B$ and the inverse relation $r'$ and compute the F1 score on $\mathcal{D}_{\leftarrow}$. To show whether the model is capable of correctly recalling the original fact $(A,r,B)$ in the first place, we additionally query the model to generate $B$ given $A$ and $r$ on $\mathcal{D}_{\rightarrow}$. For the originally pre-trained LMs without accessing Wikidata, we convert the knowledge triplets to natural language QA pairs as explained in Section~\ref{sec:PQA}.

For implementation, we select seven relation pairs $(r,r')$ from $\mathcal{D}_0$ as shown in Table~\ref{tab:onehop-template} from Appendix~\ref{sec:appx-reasoning}. For each relation, we apply the restriction that for knowledge triplet $(A,r,B)$, the inverse knowledge $(B,r',A)$ is not contained in $\mathcal{D}_0$. For each relation, we randomly sample 150 triplets from $\mathcal{D}_0$, resulting in 1,050 samples for both $\mathcal{D}_{\rightarrow}$ and $\mathcal{D}_{\leftarrow}$.

As shown in Figure~\ref{fig:T5-onehop} and \ref{fig:llama-onehop}, for all models, we can observe a limited performance increase when answering the inverse knowledge $(B,r',A)$, despite the models showing increasing memorization accuracy of the forward knowledge $(A,r,B)$. We speculate this ``no significant change'' in deduction results suggests that LMs can memorize knowledge well but are short at handling the inverse of relations.

\subsection{Compositional Reasoning} \label{sec:twohop}

We define compositional reasoning as the ability to infer $(A,r_3,C)$ given $(A,r_1,B)$ and $(B, r_2, C)$ when $(A,r_1,B) \wedge (B, r_2, C) \Rightarrow (A, r_3, C)$.
To study the model’s ability to conduct compositional reasoning over the knowledge base, we first curate a set of triplet pairs containing $(A,r_1,B)$ and $(B, r_2, C)$, denoted by $\mathcal{D}_{\wedge}=(\mathcal{D}^1_{\wedge},\mathcal{D}^2_{\wedge})$. We then form the conclusive triplet set containing $(A,r_3,C)$, denoted by $\mathcal{D}_{\Rightarrow}$. To test the model's performance, we query the model using entity $A$ and relation $r_3$, and compare the model's output with the ground-truth entity $C$ on $\mathcal{D}_{\Rightarrow}$. To show whether the model is capable of correctly recalling the conditioned facts $(A,r_1,B)$ and $(B, r_2, C)$ in the first place, we additionally query the model to generate the objects for these conditioned facts on $\mathcal{D}_{\wedge}$. Again, for the pre-trained model, we convert the knowledge triplets to natural language QA pairs.

For implementation, we formulate eight reasoning rules $r_1 \wedge r_2 \Rightarrow r_3$ of relation composition as shown in Table~\ref{tab:twohop-template} from Appendix~\ref{sec:appx-reasoning}.

For a compositional rule $(A,r_1,B) \wedge  (B,r_2,C) \Rightarrow (A, r_3, C)$, we restrict that the prior knowledge triplets $(A,r_1,B)$ and $(B,r_2,C)$ exist in the knowledge dataset while the deduction result $(A, r_3, C)$ is missing. For each reasoning rule, we randomly sample 150 examples from $\mathcal{D}_0$, resulting in 1,200 samples for both $\mathcal{D}_{\wedge}$ and $\mathcal{D}_{\Rightarrow}$. 

The results from Figure~\ref{fig:t5-twohop} and \ref{fig:llama-twohop} show that training on the KB can assist the model in performing compositional reasoning. However, there is an upper threshold; memorizing prior knowledge beyond that point may not help the model perform compositional deduction.

\section{Related Work}

\paragraph{Infusing Knowledge into LM}

Starting from seminal work by \cite{petroni-etal-2019-language} that first introduced the concept of using pre-trained language models as knowledge bases, many works investigate such viability by finetuning and evaluating the models on downstream question-answering tasks \cite{roberts-etal-2020-much,Guu2020REALMRL,moiseev-etal-2022-skill}. Notably, using salient span making \cite{moiseev-etal-2022-skill}, augmented learning objective \cite{Verga2020FactsAE}, and modifying model architecture \cite{Zhang2022GreaseLMGR, Yasunaga2022DeepBL} have shown to improve LM's performance on various open-domain question answering tasks.

When explicitly studying the capacity to store factual information, many knowledge datasets have been proposed. Among them, LAMA \cite{petroni-etal-2019-language} is based on factual and commonsense knowledge grounded to Wikipedia. WikiData5M contains 4.9M Wikidata triplets derived from \cite{wang-etal-2021-kepler}. More recently, \citet{cao-etal-2021-knowledgeable} derived WIKI-UNI with a uniform distribution of object entities, and \citet{keleg-magdy-2023-dlama} proposed DLAMA to group factual information by cultural diversity. 

Under the scope of investigating the infusing of KB into LM, \citet{bosselut-etal-2019-comet} focus on commonsense knowledge derived from ATOMIC \cite{Sap2019ATOMICAA} and ConceptNet \cite{Speer2016ConceptNet5A}, \citet{heinzerling-inui-2021-language} study the memorization capacity of BERT-based models using popular Wikidata knowledge. AutoPrompt \cite{shin-etal-2020-autoprompt} can be utilized to modify knowledge input \cite{veseli2023evaluating, veseli-etal-2023-evaluating} for triplet completion. In addition, \citet{mallen-etal-2023-trust} proposed to use retrieval-augmented LMs to help with long-tail factual knowledge.

\paragraph{Probing for Exisisting Knowledge}

Given that pre-trained LMs are sensitive to input \cite{jiang-etal-2020-know,elazar-etal-2021-measuring} and querying the model with even syntactical variations may lead to different output results \cite{longpre-etal-2021-entity}, many work have focused on the probing technique to extract knowledge stored inside LM through pre-training. For example, \citet{li-etal-2022-eliciting} study the extraction of knowledge under the setting of unsupervised knowledge-grounded conversation, \citet{alivanistos2023prompting} utilize prompt generation and post-processing techniques to probe for knowledge, while others focus on extracting specific types of factual information, such as commonsense knowledge \cite{davison-etal-2019-commonsense}, simile metaphor \cite{chen-etal-2022-probing} and biomedical facts \cite{sung-etal-2021-language}.

\section{Conclusion}
In this work, we systematically study the viability of using language models as large-scale knowledge bases. 
We propose an importance sampling algorithm to increase the efficiency of memorizing world knowledge from Wikidata. 
We investigate and evaluate three critical dimensions along this direction and conclude that large language models are able to recall a large amount of knowledge in KB through training in both fixed form as the structured KB and free form as natural language queries, with increasing flexibility when querying the world knowledge. Nevertheless, there is a significant gap between the memorization of popular knowledge and long-tail knowledge regardless of model size. 
In addition, language models, after being trained on the large-scale KB, demonstrate consistent improvement in terms of inferring new facts through some extent of reasoning. However, the amount of knowledge learned during training does not guarantee consistent improvement in reasoning capabilities, especially when it comes to inverse reasoning.
These results point to future work in utilizing language models as knowledge bases at scale, as well as further investigations on improving LMs' reasoning capability over world knowledge.

\section*{Limitations}
This work focuses on the following three aspects of treating language models as knowledge bases: memorization and accessing of knowledge base information at scale, accessing of memorized knowledge in flexible, natural language format, and inferring facts missing from the knowledge base used for training. \citet{alkhamissi2022review} proposed the following five abilities for a language model to be qualified as a knowledge base: (1) accessing of knowledge, (2) editing of knowledge, (3) consistency over semantically equivalent context, (4) reasoning over stored knowledge, (5) explainability in internal mechanisms and interoperability of outputs under a post-hoc setting. We mainly address the ability of knowledge accessing while providing a preliminary study on the reasoning ability of language models over using inverse and compositional reasoning. 
Another limitation in our work is that, due to limited computation resources, we are unable to train the models without importance sampling on the 46M triplets of $\mathcal{D}_0$. While it may require further investigation on importance sampling to improve credibility and robustness, we believe our experiments on the subset $\mathcal{D}_1$ randomly sampled from $\mathcal{D}_0$ are preliminary evidence to support our hypothesis in Section~\ref{sec:imsmp}, and serve as a good foundation for speeding up large-scale knowledge memorization.

\section*{Ethics Statement}

Large language models are known to memorize information from pre-training corpus.
Therefore, probing for stored knowledge may lead to privacy attacks against language models, such as training data extraction attacks \cite{Neel2023PrivacyII, Staab2023BeyondMV, Hartmann2023SoKMI}. For this kind of attack, an adversary can reconstruct parts of the training sample when given access to the model, leading to potential exposures of sensitive information that should not be extracted in fair and ethical usage of language models.
In addition, \citet{karamolegkou-etal-2023-copyright} confirms that language models are able to memorize a substantial portion of bestselling books with copyright that are published between 1930-2010, which demonstrates the risk of copyright violations when deploying the language models.  

For our work, the world knowledge dataset $\mathcal{D}_0$ is derived from Wikidata, which follows the CC0 (Creative Common Public Domain) Copyright\footnote{\url{https://www.wikidata.org/wiki/Wikidata:Copyright}}. In this way, we reduce the concern of language models learning sensitive or copyright information when training on the corresponding knowledge dataset.
However, we have limited control over information acquired during the pre-training of language modles. It is possible to address this issue in future work by either using language models with sensitive and copyright information removed or deploying knowledge editing methods \cite{Zhang2024ACS} to enforce data privacy and integrity.



\bibliography{anthology,custom}

\appendix

\section{Additional training and evaluation details}\label{sec:appx-model}

\paragraph{Importance Sampling with $\mathcal{D}_1$}
We train the T5-base model from its HuggingFace checkpoint\footnote{\url{https://huggingface.co/t5-base}} in \texttt{FP32} with a batch size of 300 on two NVIDIA V100 GPUs. We use the AdaFactor \cite{Shazeer2018AdafactorAL} as the optimizer with a constant learning rate of 1e-3. The evaluation batch size is 1024. We set the maximum number of training epochs to be 100 and enforce an early stopping policy to terminate the training if the model shows no improvement on the evaluation set for 10 epochs or after the exact match score on $\mathcal{D}_{1-Val}$ exceed 96\%. The model reaches the exact match threshold for early stopping for both experiments and the training time is around 2 hours and 5 hours without and without importance sampling.

\paragraph{Training on $\mathcal{D}_0$}

We train T5 models from their HuggingFace checkpoints\footnote{\url{https://huggingface.co/t5-large}} on two NVIDIA A100 GPUs, with a batch size 512 and an evaluation batch size of 1024 in \texttt{FP32} for T5-base, a batch size of 300 and an evaluation batch size of 512 in \texttt{BF16} for T5-large. We use the AdaFactor as the optimizer with a constant learning rate of 1e-3. The approximate time for one epoch of training is 15 hours for T5-base and 11 hours for T5-large. We also set the maximum number of training epochs to be 50 and enforce an early stopping policy to terminate the training if the model shows no improvement on the evaluation set for ten epochs or after the exact match score on $\mathcal{D}_{2}$ exceed 96\%. Neither model meets the early stopping criteria when training on $\mathcal{D}_0$.

We train LLaMA-2 models from their HuggingFace checkpoints\footnote{\url{https://huggingface.co/meta-llama/Llama-2-7b-hf}}\footnote{\url{https://huggingface.co/meta-llama/Llama-2-13b-hf}} on eight NVIDIA A800 GPUs in \texttt{BF16} using Deepspeed \cite{DeepSpeed} and ZeRO \cite{Rajbhandari2019ZeROMO} with Accelerate \cite{accelerate}. For LLaMA-2-7b, the training batch size is 768 and the evaluation batch size is 96; for LLaMA-2-13b, the training batch size is 400, and the evaluation batch size is 50. For both models, we use the AdamW \cite{loshchilov2019decoupled} with a constant learning rate of 1e-5 and set the maximum sequence length to 64. The approximate time for one epoch of training is 8 hours for LLaMA-2-7b and 15 hours for LLaMA-2-13b. We also set the maximum number of training epochs to be 20 and enforce an early stopping policy to terminate the training if the model shows no improvement on the evaluation set for five epochs or after the exact match score on $\mathcal{D}_2$ exceeds 96\%. Neither model meets the early stopping criteria when training on $\mathcal{D}_0$. 

\paragraph{Finetuning and Inference}

We finetune T5-base in \texttt{FP32} on two NVIDIA V100 GPUs, and T5-large in \texttt{BF16} on two NVIDIA A100 GPUs. We set the training batch size to be 256 and the evaluation batch size to be 512, with the same optimizer and learning rate as training. With a maximum epoch of 30, we enforce an early stopping policy that terminates finetuning if the model shows no improvement on the validation set for ten epochs.

For LLaMA-2 models, we perform finetuning with the same configurations as training on $\mathcal{D}_0$. However, we set the maximum number of finetuning epochs to 15 with an early stopping policy that terminates the finetuning if the model shows no improvement on the validation set for five epochs. 


\section{Reasoning rules and triplet-to-text templates for inverse and compositional reasoning} \label{sec:appx-reasoning}

In Table~\ref{tab:onehop-template}, we present the relations for inverse reasoning rule $r$ inverse of $r'$ for Section~\ref{sec:onehop}. Corresponding templates used to convert triplet with these rules to natural language question-answering dataset can be found in Table~\ref{tab:triplet2qa-onehop}.

\begin{table}[ht!]
\centering
\begin{tabular}{cc}
\hline
\multicolumn{2}{c}{Reasoning Rule: $r$ inverse of $r'$} \\
$r$                     & $r'$                     \\ \hline
sibling                    & sibling                   \\
shares border with         & shares border with        \\
father                     & child                     \\
mother                     & child                     \\
capital                    & capital of                \\
part of                    & has part                  \\
country                    & contains                  \\ \hline
\end{tabular}
\caption{Reasoning rules for inverse relations.}
\label{tab:onehop-template}
\end{table}

\begin{table}[ht!]
\centering
\begin{tabular}{MN}
\hline
$relation$  & question text                       \\ \hline
``sibling''                                                       & the sibling of $subject$ is         \\
``shares border with''                                            & $subject$ shares border with        \\
``child''                                                         & $subject$ has child                 \\
``capital of''                                                    & $subject$ is capital of             \\
``has part''                                                      & $subject$ has part                  \\
``contains''                                                      & $subject$ contains                  \\
``father''                                                        & the father of $subject$ is          \\
``mother''                                                        & the mother of $subject$ is          \\
``capital''                                                       & the capital of $subject$ is         \\
``part of''                                                       & $subject$ is part of                \\
``country''                                                       & the country $subject$ belongs to is \\ \hline
\end{tabular}
\caption{Templates for converting knowledge triplets to natural language text for Section~\ref{sec:onehop}. The first column is the $relation$ in knowledge triplet $(subject, relation, object)$ and the second column is the question text querying for $object$ using $subject$ and $relation$ in natural language.}
\label{tab:triplet2qa-onehop}
\end{table}

In Table~\ref{tab:twohop-template}, we present the relations for compositional reasoning rules $r_1 \wedge r_2 \Rightarrow r_3$ for Section~\ref{sec:twohop}. Corresponding templates used to convert triplet with these rules to natural language question-answering dataset can be found in Table~\ref{tab:triplet2qa-twohop}.

\begin{table*}[ht!]
\centering
\begin{tabular}{BXC}
\hline
\multicolumn{3}{c}{Reasoning Rule: $r_1 \wedge r_2 \Rightarrow r_3$}                                  \\
$r_1$                & $r_2$                               & $r_3$                    \\ \hline
place of birth       & country                             & country of birth         \\
place of burial      & country                             & country of burial        \\
place of publication & country                             & country of publication   \\
place of death       & country                             & country of death         \\
performer            & languages spoken, written or signed & language of work or name \\
author               & languages spoken, written or signed & language of work or name \\
father               & father                              & grandfather              \\
mother               & mother                              & grandmother              \\ \hline
\end{tabular}
\caption{Reasoning rules for relation composition.}
\label{tab:twohop-template}
\end{table*}

\begin{table*}[ht!]
    \centering
\begin{tabular}{AMc}
\hline
$r_1 \wedge r_2 \Rightarrow r_3$  & $relation$                        & question text                                           \\ \hline
\multirow{5}{*}{$r_1$}                                         & ``place of birth''                    & the place of birth of $subject$ is                      \\
                                 & ``place of burial''          & the place of burial of $subject$ is        \\
                                 & ``place of publication''     & the place of publication of $subject$ is   \\
                                 & ``place of death''           & the place of death of $subject$ is         \\
                                 & ``author''                   & the author of $subject$ is                 \\ \cline{2-3} 
\multirow{2}{*}{$r_1$ and $r_2$} & ``father''                   & the father of $subject$ is                 \\
                                 & ``mother''                   & the mother of $subject$ is                 \\ \cline{2-3} 
\multirow{2}{*}{$r_2$}           & ``country''                  & the country $subject$ belongs to is        \\
                                                               & ``langues spoken, written or signed'' & the languages spoken, written or signed by $subject$ is \\ \cline{2-3} 
\multirow{7}{*}{$r_3$}                                         & ``country of birth''                  & the country of birth of $subject$ is                    \\
                                 & ``country of burial''        & the country of burial of $subject$ is      \\
                                 & ``country of publication''   & the country of publication of $subject$ is \\
                                 & ``country of death''         & the country of death of $subject$ is       \\
                                 & ``language of work or name'' & the language of $subject$ is               \\
                                 & ``grandfather''              & the grandfather of $subject$ is            \\
                                 & ``grandmother''              & the grandmother of $subject$ is            \\ \hline
\end{tabular}
    \caption{Templates for converting knowledge triplets to natural language text for Section~\ref{sec:twohop}.  The first column indicates where $relation$ appears in compositional reasoning $r_1 \wedge r_2 \Rightarrow r_3$, the second column is the $relation$ in knowledge triplet $(subject, relation, object)$, and the third column is the question text querying for $object$ using $subject$ and $relation$ in natural language.}
    \label{tab:triplet2qa-twohop}
\end{table*}

\section{Dataset and open-source projects}

In preparing our own world knowledge dataset $\mathcal{D}_0$ of scale similar to latest KBs, we use the CC0-licensed English Wikidata \cite{Tanon2016WikiData} as the source of world knowledge and an MIT-licensed code project released by \citet{philipp2021wikidatacore} to filter away knowledge irrelevant to common linguistic tasks. We further derive various subsets from $\mathcal{D}_0$ to study the memorization behavior of language models as in Section~\ref{sec:imsmp}, \ref{sec:KB}, \ref{sec:onehop} and \ref{sec:twohop}.

Our experiments on free-form information in Section~\ref{sec:PQA} are based on the PopQA dataset released by \citet{mallen-etal-2023-trust} under MIT License. For general missing fact completion in Section~\ref{sec:missing-facts}, we utilize the portion of human-annotated missing facts from the dataset created by \citet{veseli2023evaluating}, which is open-sourced into a public repository.

For experiments on LLaMA-2 models, we also employ a public fork of HuggingFace transformer library to address the left padding problem that may impact the inference results. The fork is licensed under Apache-2.0 and hosted on \url{https://github.com/yizhongw/transformers/tree/left_padding}.

\end{document}